%% file: mosaic.tex
\documentclass{article}

\usepackage[preprint,nonatbib]{neurips_2022}

\usepackage[utf8]{inputenc} 
\usepackage[T1]{fontenc}    
\usepackage{hyperref}       
\usepackage{url}            
\usepackage{booktabs}       
\usepackage{amsfonts}       
\usepackage{nicefrac}       
\usepackage{microtype}      
\usepackage{xcolor}         
\usepackage{amsmath}        
\usepackage{graphicx}       
\usepackage{multirow}
\usepackage{subcaption}
\usepackage{tikz}

\def \etal {\emph{et al.}}

\title{MOSAIC: Masked Optimisation with Selective Attention for Image Reconstruction}

\author{\vspace{-5mm}\\
Pamuditha Somarathne$^{1}$, Tharindu Wickremasinghe$^{1}$, Amashi Niwarthana$^{1}$, A. Thieshanthan$^{1}$,\\ 
\vspace{2mm}
Chamira U.S. Edussooriya$^{1}$, and Dushan N. Wadduwage$^{2,*}$\\
$^{1}$ Department of Electronic and Telecommunications, Engineering, University of Moratuwa, Sri Lanka\\
\vspace{2mm}
$^{2}$ Center for Advanced Imaging, Harvard University, Cambridge, Massachusetts\\
$^{*}$ \texttt{wadduwage@fas.harvard.edu}\\  
}

\begin{document}
\maketitle

\begin{abstract}
Compressive sensing (CS) reconstructs images from sub-Nyquist measurements by solving a sparsity-regularized inverse problem. Traditional CS solvers use iterative optimizers with hand crafted sparsifiers, while early data-driven methods directly learn an inverse mapping from the low-dimensional measurement space to the original image space. The latter outperforms the former, but is restrictive to a pre-defined measurement domain. More recent, deep unrolling methods combine traditional proximal gradient methods and data-driven approaches to iteratively refine an image approximation. To achieve higher accuracy, it has also been suggested to learn both the sampling matrix, and the choice of measurement vectors adaptively. Contrary to the current trend, in this work we hypothesize that a general inverse mapping from a random set of compressed measurements to the image domain exists for a given measurement basis, and can be learned. Such a model is single-shot, non-restrictive and does not parametrize the sampling process. To this end, we propose MOSAIC, a novel compressive sensing framework to reconstruct images given any random selection of measurements, sampled using a fixed basis. Motivated by the uneven distribution of information across measurements, MOSAIC incorporates an embedding technique to efficiently apply attention mechanisms on an encoded sequence of measurements, while dispensing the need to use unrolled deep networks. A range of experiments validate our proposed architecture as a promising alternative for existing CS reconstruction methods, by achieving the state-of-the-art for metrics of reconstruction accuracy on standard datasets.

\end{abstract}

\input{Sections/1_Introduction}

\input{Sections/2_Related_work}
\input{Sections/3_Method}
\input{Sections/4_Experiments_and_Results}
\input{Sections/5_Discussion}

\input{Sections/6_Conclusion}

\newpage
\input{Sections/7_Supplementary}

\newpage
\bibliographystyle{plain}
\bibliography{bibliography}

\end{document}

%% file: Sections/1_Introduction.tex
\section{Introduction}

Compressive sensing (CS) has advanced signal processing and data acquisition by enabling accurate signal recovery from highly under-sampled measurements, with numerous applications in imaging, speech/audio processing, and communication systems~\cite{rani2018,qin2018}. The typical CS formalism for imaging employs a sampling matrix, $\mathbf{A}_{m\times n}$, to compressively measure an image signal $\mathbf{x} \in \mathbb{R}^{n}$ as a linear projection $\mathbf{y} = \mathbf{A} \mathbf{x} (m \ll n)$. Given $\mathbf{y} \in \mathbb{R}^{m}$, the subsequent reconstruction model should solve an ill-posed inverse problem to reconstruct the image $\mathbf{\hat{x}} \in \mathbb{R}^{n}$.
The seminal works of Donoho~\cite{csdonoho} and Candes~\etal~\cite{cscandes} inspired reconstruction algorithms~\cite{ihta_blumensath,ista_cai2009iterative,amp_donoho2010approximate} with theoretical guarantees. These methods employ a structural prior $\mathcal{R}(.)$ with a regularization parameter $\lambda$ to form an optimization objective:
\begin{equation}
    \underset{\mathbf{x}}{\operatorname{argmin}}\: ||\mathbf{A} \mathbf{x} - y||_2^2 + \lambda \mathcal{R}(\mathbf{x}).
    \label{eq:CSbasic}
\end{equation}
Consequently, projected gradient methods~\cite{elad2006imageprojgrad} have been widely explored which iteratively approximate better reconstructions. These algorithms have a gradient update step to form a residual $\mathbf{\hat{x}}_{k} - \alpha \nabla f(\mathbf{\hat{x}}_k)$, after which the updated residual is projected back on to a convex set including $\mathbf{x}$. 
\begin{equation}
\hat{\mathbf{x}}_{k+1} = \operatorname{P}_{\lambda\mathcal{R}}\left(\hat{\mathbf{x}}_k - \alpha \nabla f(\mathbf{\hat{x}}_k)\right) ; \quad f(\mathbf{\hat{x}}) = ||\mathbf{A} \mathbf{\hat{x}} - y||_2^2,
\label{eq:projectedbasic}
\end{equation}
Here $\hat{\mathbf{x}}_{k}$ and $\hat{\mathbf{x}}_{k+1}$ are the signal estimates at iterations $k$ and $k+1$, respectively. $\operatorname{P}_{\lambda\mathcal{R}}$ is the projection operator, parametrized by $\lambda$ and the choice of $\mathcal{R}$. 
Inevitably, these "traditional" iterative algorithms were replaced with deep learning models to achieve faster inference and higher accuracy by parametrizing terms of equation~\eqref{eq:projectedbasic} into learnable weights with a limited number of iterations. They achieve state-of-the-art reconstruction accuracy for standard image datasets, while having the added advantage of requiring only a single forward pass step at inference.
In line with exploiting priors of natural images, it is worth noting that natural images have well-exploited structures in some spectral decompositions such as in the frequency domain ~\cite{VANDERSCHAAF19962759, wilson2008}. This inherent structure creates a pattern, with each frequency carrying a different information content. We could extend this idea to compressive sensing, when each measurement vector (row $A_{i}$) of the sampling matrix $\mathbf{A}$ is structured such that it corresponds to a different basis frequency that spans the image space.
Therefore, each measurement would correspond to sampling at a certain frequency, and given $\mathbf{y}_{m \times 1}$, it is possible to consider $(y_{i}, A_{i})$ pairs for $1 \leq i \leq m$, as known priors for the reconstruction objective.
However, structural priors may not be sufficient to solve an ill-posed inverse problem with high accuracy.

An interesting way to approach CS reconstruction, is to convert it to a `measurement filling problem' similar to that of sentence completion in natural language processing (NLP)~\cite{devlin2018bert}. Drawing intuitions from attention-based models in NLP, a potential approach is to employ attention in the measurement space, so that more salient measurements are given more attention. Nevertheless, a crucial challenge is to identify and encode the maximum amount of information from any random set of measurements. While significant progress has been made in compressive sensing, we find that many data driven methods parametrize the sampling process, restricting it to a learnt $\mathbf{A}$.

In this context, we present an attention-based framework for applying CS-specific inductive biases for a deep CS model, for a fixed measurement basis. Specifically, our contributions are as follows:
\begin{enumerate}
     \item We propose a novel approach for reconstruction of compressively sensed images, exploiting the structure of an orthogonal sampling basis, while dispensing the need for deep unrolling.
     
     \item We present "MOSAIC"; \textbf{M}asked \textbf{O}ptimisation with \textbf{S}elective
\textbf{A}ttention for \textbf{I}mage Re\textbf{c}onstruction; an end-to-end trainable deep reconstruction model that is not restricted to a learnt sampling process.  
     
     \item We demonstrate the effectiveness of "MOSAIC" on compressive sensing, achieving significant improvements in PSNR and SSIM over previous state-of-the-art methods on benchmark datasets, as well as a qualitative review.  
 \end{enumerate}

%% file: Sections/2_Related_work.tex
\section{Related Work}

\subsection{Projection-based Deep Unrolling}
The first family of deep CS reconstruction used black-box CNN models~\cite{kulkarni2016reconnet, shi2019imageCNN, shi2019scalableCSCNN}. However, these were lacking interpretability, and were restricted to the specific learnt sampling matrix.
Inspired by sparse dictionary encoding algorithms such as Iterative Shrinkage and Thresholding Algorithm (ISTA)~\cite{daubechies2004ista}, the next generation of deep models unroll iterative algorithms such as equation \eqref{eq:projectedbasic} to a series of blocks~\cite {istanet2018, ldamp2017, admmcsnet2020}. To various degrees, these `deep unrolling' models rely on the guarantees of projected gradient methods to form their inductive bias while striving for some level of interpretability. This translates to finding a sweet spot between an inductive bias, and model flexibility, by the conversion of certain parameters to learnable weights.
The latest approaches~\cite{istanetplus2021, zhang2020ampnet, shen2022transcsnet, chen2022fsoinet}, however, only use CS recovery algorithms as inspiration for the model architecture. For example, ISTA-Net++~\cite{istanetplus2021}, uses CNNs, while the more recent TransCS~\cite{shen2022transcsnet} uses a hybrid of transformers and CNNs to model the parametrized projection and gradient update step. As an underlying pattern, these models learn an embedding from an image, update a gradient step in the embedding-space, and project back to image-space resembling a deep autoencoder~\cite{hinton2006reducingdimsAE}. The reconstruction path is a stack of such autoencoders, progressively refining/denoising $\mathbf{\hat{x}}$, with $\Phi$ and $\mathbf{y}$ acting as priors.
However, we notice that these methods update model weights through the gradient of the residual norm $ ||\Phi \mathbf{\hat{x}} - \mathbf{y}||_2$, and that there is no separate treatment for each measurement $\mathbf{y}_{i}$ of $\mathbf{y}$. 

\subsection{Attention for Image Reconstruction}

Masked image modeling is a technique that utilizes images that have been corrupted by masking to learn useful representations. Pioneered with stacked autoencoders \cite{JMLR:v11:vincent10a} using convolutional neural networks (CNNs), while numerous methods \cite{bao2022beit, he2022mae, xie2021simmim} have employed the mask-word approach derived from natural language processing (NLP) to randomly obscure image patches within the spatial domain through the utilization of vision transformers \cite{visiontransformer2021an, liu2021Swin}. Recent methods incorporating masked autoencoders~\cite{he2022mae}, have shown great promise in similar image processing tasks~\cite{NEURIPS2022_ea370419, NEURIPS2022_e97d1081}, where they efficiently extract spatial features that are representative of the original data. Learning through autoencoders is a challenge in CS, since the measurement $\mathbf{y}$ itself is low dimensional. Therefore, we propose a structured method of expanding the dimensionality of $\mathbf{y}$, that would enable masked learning in an embedding-space.

\subsection{Representation Learning from Transformed Domains}
Studies on how image details are represented in frequency space have been extensively studied~\cite{oppenheim1979phase, oppenheim1981importance}, to show that the phase component of a signal in the frequency domain captures high-level semantics, while the amplitude component preserves low-level statistics. This has motivated deep learning methods to incorporate methods to use transformations such as discrete cosine transform (DCT)~\cite{li2023dCTFormer} and discrete Fourier transform (DFT)~\cite{rao2021globalFFT} to learn image features for subsequent downstream tasks. With the success of masked approaches for visual pretraining, \cite{xie2023maskedfreqmodel} employs masked learning in DFT domain, to learn image features in a self-supervised setting. However, as the transformations in CS are ill-posed, the challenge is to solve for an inverse transformation. Therefore, learning features in the measured space is not as straightforward as existing approaches. Furthermore, the existence on the spectral bias of neural networks towards low frequency information~\cite{pmlr-v97-rahaman19a, dziedzic2019bandspectra} has motivated careful design of deep networks for tasks that are more reliant on high frequency information~\cite{mildenhall2021nerf}. Drawing parallels onto the measurement domain, we address these concerns with a measurement embedding scheme, which facilitates masked learning through a mapping into higher dimensions.

%% file: Sections/3_Method.tex
\input{Images/fig2}

\section{Method}

Let $\mathbf{X}_{\sqrt n\times\sqrt n}$  be the image signal, with pixels values $x_{i,j}\in \mathbb{R}$, at pixel locations  $\left(i,j\right)\in S=\left\{\left(i^\prime,j^\prime\right)\in \mathbb{Z}^2\ \right|\ 1\le i\prime,j\prime\le\sqrt n \}$. Here $n = \|\mathbf{X}\|_0$ is the number of pixels in the image. Select the transformation matrix $\Phi_{\sqrt n\times\sqrt n}$ such that a well-posed measurement $\mathbf{Y}_{\sqrt n\times\sqrt n}$ with entries $y_{i,j}\in \mathbb{R}$ is given by \eqref{eq:samplingbasic}, and depicted in Figure~\ref{fig:sampling_diagram}.

\begin{equation}
 \mathbf{Y}_{\sqrt{n} \times \sqrt{n}} = (\Phi \thickspace \mathbf{X} \thickspace \Phi^T )/k; \quad k \mathbf{I}_{n} = \Phi \Phi^T.
 \label{eq:samplingbasic}
\end{equation}

\input{Images/fig3}

From equation~\eqref{eq:samplingbasic},  $\forall \left(i,j\right) \in S $, for $\mathbf{\phi_{i}},\mathbf{\phi_{j}}$ rows of $\mathbf{\Phi}$, we can write 
\begin{equation}
 y_{i,j} = \frac{\mathbf{\phi_{i}} \thickspace \mathbf{X} \thickspace \mathbf{\phi}^T_{j} }{k}. 
 \label{eq:samplingbasic2}
\end{equation}

Furthermore, note the resemblance of the following decomposition, to a spectral decomposition of $\mathbf{X}$, where a one-to-one mapping exists between the choice of measurement basis vectors $(\mathbf{\phi_{i}},\phi_{j})$, and the measurement value $y_{i,j}$.
\begin{equation}
 \mathbf{X} = k \ \sum_{i=1}^{\sqrt{n}} \sum_{j=1}^{\sqrt{n}} \mathbf{\phi_{i}}^T y_{i,j} \mathbf{\phi_{j}} .
 \label{eq:samplingbasicspectral}
\end{equation}

\textbf{Compressive sensing:}
As shown in equation \eqref{eq:samplingbasic2}, all measurements are indexed by $\left(i,j\right)$ pairs and the set of all measurement indices $S$ has a cardinality,  $\left|S\right|=n$. To compressively sample $\mathbf{X}$ we randomly select a subset of $m$ measurements with indices in $S_{C}\subset S \text{ where} \left|S_C\right| = m $, following a uniform distribution. The uniform distribution ensures that each measurement has equal weight and minimizes potential sampling biases. Now we can define the compression factor $\gamma$ as,
\begin{equation}
    \gamma = \frac{m}{n} < 1.
    \label{eq:gammadefinition}
\end{equation}

\textbf{Embedding:}
Next, using the structure of $\Phi$, we embed each measurement  $y_{i,j}$, for $\left(i,j\right)\in S_C$, as follows,
\begin{equation}
    \mathbf{E}_{i,j} = \mathbf{\phi_{i}}^T y_{i,j} \mathbf{\phi_{j}}.
    \label{eq:Eij}
\end{equation}

This facilitates the decomposition of the measurements, generating a high dimensional representation of the sampled measurements that can be fed to the subsequent model. Having a high dimensional input alleviates possible spectral biases of the model as motivated by \cite{pmlr-v97-rahaman19a}.

Each $E_{i,j}$ is then further embedded using a learnable embedding $T(.)$ and positional encoded using a positional embedding $Pos\left(.\right)$ to get the final embedding $\mathbf{z}_{i,j}$ as,
\begin{equation}
    \mathbf{z}_{i,j}  = f(\mathbf{E}_{i,j}, (i,j)) = T(\mathbf{E}_{i,j}) + Pos(i,j).
\end{equation}
 
\textbf{Masked encoder:}
Similar to the original MAE work~\cite{he2022mae} our positional embedding,  $Pos\left(.\right)$ , preserves the original location of the measurement $y_{i,j}$ in $\mathbf{Y}_{\sqrt n\times\sqrt n}$. Thus, we think about the sampling process as a “masking” operation on $\mathbf{Y}_{\sqrt n\times\sqrt n}$. Here, $y_{i,j}$’s with $\left(i,j\right)\in S_{C}$ are unmasked measurements while ones with $\left(i,j\right)\in {S-S_{C}}$ are masked. 

Next, we treat $\mathbf{z}_{i,j}$’s as an $m$ element sequence, $\mathbf{z}$, such that a given $\mathbf{z}_{i,j}$’s index $k\in\left\{k^\prime\in \mathbf{Z} \right|\ 1\le k^\prime \le m \}$ in the sequence is given by the sorting operator on $S_{C}$ as,
\begin{equation}
    k = sort\left(\ \left(i,j\right)\ \right|\ S_C\ ).
\end{equation}
Then the sequence $\mathbf{z}$ can be written as, 
\begin{equation}
    \mathbf{z} = (\mathbf{z_1},\mathbf{z_2},...,\mathbf{z_k},...,\mathbf{z_m}),
\end{equation}
where $\mathbf{z_{i,j}}$ is the same as $\mathbf{z}_{sort\left(\ \left(i,j\right)\ \right|\ S_C\ )}$.

We employ an asymmetric design as outlined in Figure~\ref{fig:reconstruction_diagram} inspired by \cite{he2022mae} in that it differs from classical encoder-decoder designs. Specifically, the encoder $g(.)$ operates on a sequence of measurement latents ${\mathbf{z}_k}$'s, while the decoder aims to reconstruct the full image that was measured, in the pixel domain.
As such, we encode $\mathbf{z}$ using the learnable encoder $g(.)$, to get $\mathbf{z}_{enc} = g(\mathbf{z})$.\\

\textbf{Unmasked decoder:}
Following the asymmetric design, our decoder, $h(.)$, takes an $n$ element sequence $ \mathbf{z}_{um}$  as its input. The indices, of $\mathbf{z}_{um}$ are related to the pixel locations ${\left(i,j\right)}$ , of the to-be reconstructed image, ${\hat{X}}_{\sqrt n\times\sqrt n}$ through some un-flattening operator, such that,
\begin{equation}
    \left(i,j\right)= unflatten\left(l\ \right|\ shape(\hat{X})).
\end{equation}

Here $l \in\left\{l^{\prime}\in Z\right|\ 1\le l^{\prime}\le n\} \text{ and } shape(\hat{X})= (\sqrt n ,\sqrt n)$ is the shape of the image. We form the `unmasked' latents $ \mathbf{z}_{um}$, by "filling the gaps" of the embeddings that correspond to masked out measurements as,
\begin{equation}
    \mathbf{z_{um}} = (\mathbf{z}_{um,1},\mathbf{z}_{um,2},...,\mathbf{z}_{um,l},...,\mathbf{z}_{um,n}),
\end{equation}
where
\begin{equation}
    \mathbf{z}_{um, \ l}  =
    \begin{cases}
        \mathbf{z}_{enc,sort(l)}, &\text{if }l \in S_{C}\\
        \mathbf{p}, &\text{otherwise}.
    \end{cases}
\end{equation}

In the above equation $\mathbf{z}_{um,\ l}$ is the $l^{th}$ element of $ \mathbf{z}_{um}$,  $\mathbf{z}_{(enc,\ \ k)}$ is the $k^{th}$ element in $\mathbf{z}_{enc}$ with $k = sort(unflatten(l)|S_C)$ and $\mathbf{p}$ is a learned placeholder. We then decode $\mathbf{z}_{in}$ using the decoder $h(.)$,
\begin{equation}
    \hat{\mathbf{x}} = h(\mathbf{z}_{um})
\end{equation}
and unflattened the output to get a 2D prediction $\mathbf{\hat{X}}$.
The reconstructed sequence  $\mathbf{\hat{X}}$ has the same indexing as $\mathbf{z}_{in}$, and we get the final reconstructed image $\mathbf{\hat{X}}_{\sqrt n\times\sqrt n}$ with pixels $\hat{x}_{i,j}$ where,

\begin{equation}
    \hat{\mathbf{x}}_{i,j} = \hat{\mathbf{x}}_{unflatten(l)} = \hat{\mathbf{x}}_l.
\end{equation}

%% file: Images/fig2.tex
\begin{figure}
  \centering
  \includegraphics[width=\linewidth]
  {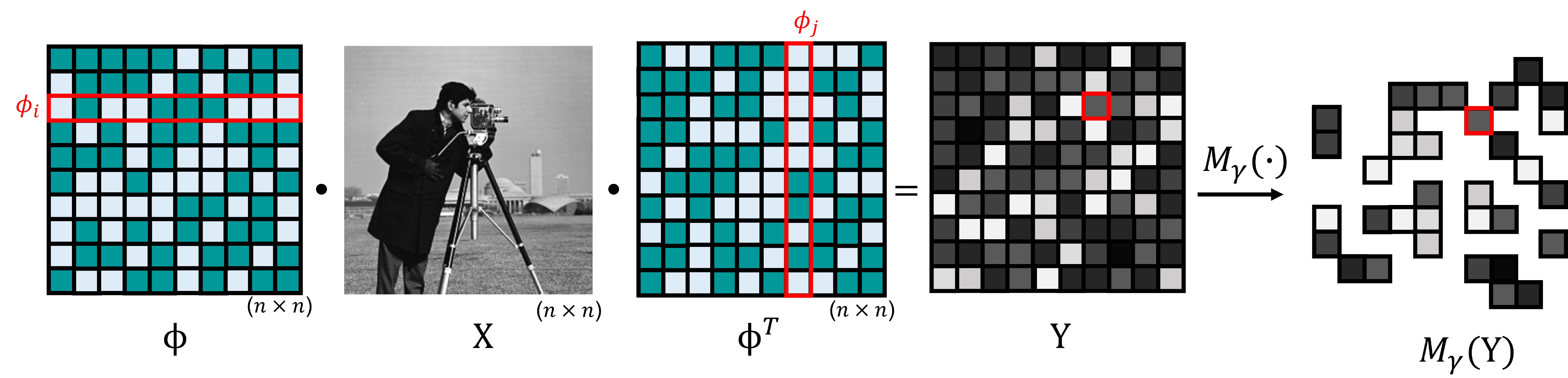}
  \caption{The proposed sampling scheme for compressive sensing of a 2D image: A uniform random sampling from a set of possible measurements $\mathbf{Y}$, is modelled as a uniform random masking $\mathit{M}$ with a compression factor $\gamma$ }
  \label{fig:sampling_diagram}
\end{figure}

%% file: Images/fig3.tex
\begin{figure}
  \centering
  \includegraphics[width=\linewidth]
  {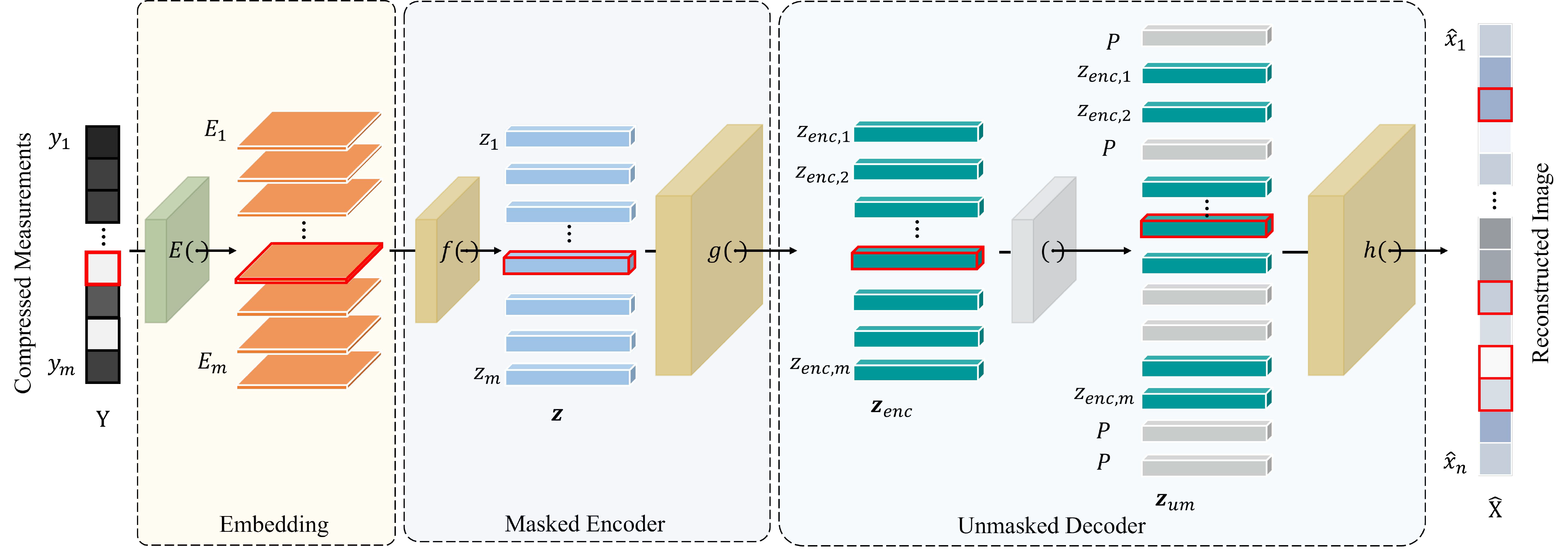}
  \caption{The proposed pipeline for reconstruction: Each sampled measurement $y_{i,j}$ is first projected to a higher dimensional $\mathbf{E}_{i,j}$ by exploiting the structure of sampling matrix $\Phi$. The masked encoder $g(.)$ maps these embeddings which are treated as a sequence of length $m$. In the decoder $h(.)$, the masked embeddings $\mathbf{z}_{enc}$ are padded with placeholer $\mathbf{p}$, and transformed to predict $\hat{\mathbf{x}} \in \mathbb{R}^n$. Here, $f(.)$ encodes the handcrafted embeddings into latents $\mathbf{z}$, while $g(.)$ and $h(.)$ are self-attention layers.}
  \label{fig:reconstruction_diagram}
\end{figure}

%% file: Sections/4_Experiments_and_Results.tex
\section{Experiments and Results}
\subsection{Experimental Setup}
\textbf{Datasets:} 
We use the standard CIFAR-10~\cite{Krizhevsky2009LearningMLCIFAR10} training set (50000 images) for training the model, and use the training set of BSD500 dataset (401 images)~\cite{bsd100} for fine tuning. The test results are acquired using three widely used benchmark datasets: Set11~\cite{kulkarni2016reconnet}, BSD100~\cite{bsd100} and Urban100~\cite{Huang-CVPR-2015}. 

Following an image patch-training approach~\cite{kulkarni2016reconnet}, MOSAIC is designed for an image patch $\mathbf{X}_{n \times n} \text{with } n = 32$. Following~\cite{istanet2018,istanetplus2021,shen2022transcsnet}, we convert 3-channel (RGB) images into YCrCb format, and only pass the luma (Y) component to train and validate the model. However, we do not perform any augmentations on the training images during pre-processing.

\textbf{Metrics:}
To be faithfully comparable with previous work~\cite{istanet2018,chen2022fsoinet,zhang2020ampnet,shen2022transcsnet}, we use structured similarity index metric (SSIM) and peak signal to noise ratio (PSNR) as evaluation metrics.

\textbf{Baseline:}
Considering the ease of replicability in real world applications, we set $\Phi=\mathcal{H}_{32\times32}$ where $\mathcal{H}$ is the Walsh-Hadamard matrix. The encoder $g_{16}(.)$ contains $16$ transformer blocks with $2$ heads with the transformer embedding  $\mathbf{z}_k \in \mathbb{R}^{32}$. The decoder $h_{16}(.)$ follows the same structure and embedding depth as the encoder. The combined model is trained on mean square error loss, as $\mathcal{L}_{mse} (\mathbf{x}, \mathbf{\hat{x}})$. 

\textbf{Training:} For each compression ratio, the model is trained for 500 epochs on CIFAR-10 and fine-tuned on BSD500 for 200 epochs. The training process starts with a $1\times10^{-3}$ learning rate and an exponential learning rate decay of $\tau = 0.9$ at each $15000$ steps. The fine-tuning process starts with a learning rate of $1\times10^{-4}$ and follows the same learning rate decay. The Adam optimizer~\cite{kingma2014adam} is used in both training and fine-tuning stages.

\subsection{Quantitative Results}

Table~\ref{tab:ssim-psnr-comp} contains the SSIM and PSNR metrics of our model compared with published performance of recent models. We compare the model outputs at five compression ratios that have been commonly reported in recent work. 
To maintain fair comparison, we only recreate results (marked with *) for models with publicly available model weights applicable to each  $\gamma$, and do not report values (-) for models which require $\gamma$-specific training. Please refer to the supplementary for a more extensive report.

MOSAIC was able to out-perform previous models across almost all of the compression ratios. This shows that faithful reconstructions are consistently possible, through our proposed embedding and self attention scheme.
It is worth noting that MOSAIC performed well across the range of compression factors, even without a learnt sampling matrix $\Phi$. It clearly outperformed CS methods which incorporate a fixed sampling matrix~\cite{istanetplus2021, song2023dpcdun}, and exceeded almost all models with learnt-sampling (LS) matrix as well. Therefore, MOSAIC justifies using a predefined, structured sampling matrix, and demonstrates the advantage of using the known structure as a prior for reconstruction. Moreover, the use of sampling matrix in this study is not restricted to or trained on any particular data distribution, thus obviating concerns regarding its generalizability.

\input{tables/results_main.tex}

\textbf{Parameter count:} Transformer models are notorious for their large computational cost. Therefore, we analyse the number of learnable parameters of MOSAIC in Table~\ref{tab:ablation} to see if the performance gain is justifiable compared to the required parameters.

Furthermore, we compare our model with existing work to compare how our model performs in the parameter-performance trade-off in Figure~\ref{fig:psnrvsparam}.
Specifically, in comparison to TransCS~\cite{shen2022transcsnet}, which is the most recent CS work with transformers, our model is shown to be significantly light-weight.
\input{Images/psnr_vs_param}

\textbf{Robustness to noise:}
Compressive sensing systems almost always deal with some level of noise during image acquisition~\cite{arias2011noise}. Therefore, it is prudent to verify the reliability of our model when the measurements are corrupted with different degrees of noise.
To this end, we follow~\cite{shen2022transcsnet}, by adding Gaussian noise with different standard deviations ($\sigma$) to the input before inference on the BSD100 dataset. 
Table~\ref{table:noise} illustrates that our model is more robust than previous work, even-though we do not augment our images with distortions in our training pipeline. It can be noted that MOSAIC compares better as $\gamma$ increases, with larger margins. This experiment also validates the generalizability of our model, and its ability to better isolate and reconstruct the underlying image pattern, distinguishing from random noise.
\input{tables/noise_robust.tex}

\subsection{Qualitative Results}
\input{Images/models_comparison}

A visual analysis of the test reconstructions are provided to further validate the quantitative results. Compared to recent deep CS reconstruction tasks, it can be observed that visual level discrepancies become obvious only at low compression ratios. 
As an example, the Zebra image from Set14 dataset~\cite{Zeyde2010set14} is compared in Figure~\ref{fig:modelcomparison}, which shows that the detail in both the foreground and background are harder to capture.

\input{Images/meta_comparison}
Since MOSAIC does not rely on convolutional layers to refine a predicted image, it is less likely to filter out high frequency content. This can be observed in the butterfly image from Set11~\cite{kulkarni2016reconnet} in Figure~\ref{fig:modelcomparison}, where the small white patches, and fine grained detail are lost in all previous work, while it is preserved in our reconstruction.

We have compared only across the luminance channel since it is the only channel that is compared quantitatively among related literature ~\cite{istanet2018,istanetplus2021,chen2022casnet}.

\subsection{Ablation Study}
We ablate the effectiveness of different modules of MOSAIC model to identify the importance of each of them. To be easily comparable across ablations, we use $\gamma=0.25$.

\textbf{Effect of transformer block count:} We change the transformer block count in the encoder and decoder parts of the reconstruction module. Evaluation metrics are shown in Table~\ref{tab:ablation}. It is observed that the model output shows only a slight variation for most of the larger configurations. Therefore, we select the 16-layer configuration as our baseline model.

\input{tables/ablation.tex}

\textbf{Effect of embedding $\mathbf{E}$:}\\
We replace $\mathbf{E}_{i,j}$ of equation \ref{eq:samplingbasic2} with a matrix of all ones, to evaluate the effect of embedding the sampling matrix information ($\phi_{i}, \phi_{j}$ ) into the measurement $\mathbf{y}_{i,j}$ before being input to the transformer. 
The ablation results on test sets are shown on Table~\ref{tab:withoutembedding}. The performance without $\mathbf{E}_{i,j}$ dropped by 50\% for PSNR, and more than 60\% for SSIM. 
It is clear that our intuitions of decomposing a sample $\mathbf{X}$ as in equation~\ref{eq:samplingbasicspectral} resulted in an structured embedding, which is critical for accurate reconstruction. Even-though a sufficiently parametrized deep learning model should in theory be able to learn such decompositions, the strong inductive bias given by explicitly using $\phi_{i}^T\phi_{j}$ has shown increase performance significantly.

%% file: tables/results_main.tex
\begin{table}
  \caption{PSNR/SSIM performance comparison of CS methods with different CS ratios on the Set11, BSD100 and Urban100 datasets. The best results are highlighted. LS stands for methods that use learnt sampling matrices, and * denotes results recreated using publicly available model weights.}
  \label{tab:ssim-psnr-comp}
  \centering
  \resizebox{\textwidth}{!}{
  \begin{tabular}{clcccccc}
    \toprule
    & \multirow{2}{*}{Method} & \multirow{2}{*}{LS} & \multicolumn{5}{c}{PSNR/SSIM for sampling ratios}\\
    \cmidrule{4-8}
     & & & 0.01 & 0.04 & 0.1 & 0.25 & 0.5 \\
    \toprule

    \multirow{5}{*}{\rotatebox[origin=c]{90}{Set11}} & ISTA-Net++~\cite{istanetplus2021} & \texttimes & 18.29/0.459 & 22.91/0.679 & 27.54/0.835 & 32.70/0.927 & 37.47/0.967 \\
     & AMP-Net~\cite{zhang2020ampnet} & \checkmark & 20.20/0.558 & 25.26/0.772 & 29.40/0.878 & 34.63/0.948 & 40.33/0.980 \\
     & FSOINet~\cite{chen2022fsoinet} & \checkmark & 21.73/0.594 & 26.37/0.812* & \textbf{30.44}/0.902 & 35.80/0.960* & 41.08/0.983 \\
     & CASNet~\cite{chen2022casnet} & \checkmark & 21.97/0.614 & 26.41/0.815 & 30.36/0.901 & 35.67/0.959 & 40.93/0.983 \\
     & DPC-DUN~\cite{song2023dpcdun} & \texttimes & - & - & 29.40/0.880 & 34.69/0.948 & 39.84/0.978 \\
     & \textbf{MOSAIC (Ours)} & \texttimes & \textbf{23.68}/\textbf{0.710} & \textbf{28.28}/\textbf{0.871} & 29.93/\textbf{0.905} & \textbf{36.02}/\textbf{0.970} & \textbf{45.46}/\textbf{0.997} \\ \midrule

    \multirow{5}{*}{\rotatebox[origin=c]{90}{BSD100}} & ISTA-Net++~\cite{istanetplus2021} & \checkmark & 19.98/0.449 & 23.26/0.587 & 25.86/0.717 & 29.48/0.854 & 33.97/0.941 \\
     & AMP-Net~\cite{zhang2020ampnet} & \checkmark & 23.45/0.554 & 26.35/0.681 & 28.87/0.792 & 32.69/0.903 & 37.71/0.967 \\
     & CSNet+~\cite{shi2019imageCNN} & \checkmark & - & 25.79/0.674 & 28.29/0.796 & 31.50/0.899  & 36.35/0.964 \\
     & TransCS~\cite{shen2022transcsnet} & \checkmark & - & 26.25/0.689 & 28.79/0.806 & 32.66/0.910  & 37.70/0.969 \\
     & \textbf{MOSAIC (Ours)} & \texttimes & \textbf{24.59}/\textbf{0.577} & \textbf{28.12}/\textbf{0.832} & \textbf{29.33}/\textbf{0.866} & \textbf{34.63}/\textbf{0.955} & \textbf{45.07}/\textbf{0.996} \\ \midrule

     \multirow{6}{*}{\rotatebox[origin=c]{90}{Urban100}} & ISTA-Net++\cite{istanetplus2021} & \checkmark& 17.48/0.416 & 20.96/0.599 & 24.66/0.762 & 29.51/0.894 & 34.41/0.957 \\
     & AMP-Net~\cite{zhang2020ampnet} & \checkmark & 20.90/0.533 & 24.15/0.703 & 27.38/0.827 & 32.19/0.926 & 37.51/0.973 \\
     & TransCS~\cite{shen2022transcsnet} & \checkmark & - & 23.23/0.702 & 26.73/0.842 & 31.72/0.933  & 37.20/0.976 \\
     & FSOINet~\cite{chen2022fsoinet} & \checkmark & 19.87/0.522 & 22.71/0.704* & \textbf{27.53}/0.863 & 32.17/0.950* & 37.80/0.978 \\
     & DPC-DUN~\cite{song2023dpcdun} & \texttimes & - & - & 26.99/0.835 & 32.36/0.932 & 37.52/0.974 \\
     & \textbf{MOSAIC (Ours)} & \texttimes & \textbf{21.35}/\textbf{0.571} & \textbf{25.20}/\textbf{0.830} & 26.72/\textbf{0.875} & \textbf{32.43}/\textbf{0.961} & \textbf{42.41}/\textbf{0.996} \\
    \bottomrule
  \end{tabular}}
\end{table}

%% file: Images/psnr_vs_param.tex
\begin{figure}
  \centering
  \includegraphics[width=0.4\linewidth]{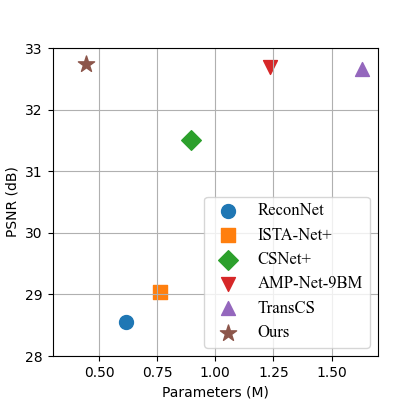}
  \caption{Relationship Between Model Parameters and Performance in PSNR. Comparison is on the BSD100 dataset at a compression ratio of 0.25}
  \label{fig:psnrvsparam}
\end{figure}

%% file: tables/noise_robust.tex
\begin{table}[!t]
\vspace{-10pt}
\centering
\caption{Evaluation of robustness to noise during inference on BSD100 dataset.}
\label{table:noise}
\resizebox{\textwidth}{!}{
\begin{tabular}{ccccccc}
\hline
$\sigma$ & $\gamma$ & ISTA-Net$^+$ ~\cite{istanetplus2021} & CSNet$^+$~\cite{shi2019imageCNN} & AMP-Net-9BM~\cite{zhang2020ampnet} & TransCS~\cite{shen2022transcsnet} & \textbf{MOSAIC (Ours)} \\ 
\hline
0.001 & 0.04 & 19.50/0.4401 & 25.18/0.6503 & 24.89/0.6461 & 26.11/0.6790 & \textbf{28.22/0.8337} \\
0.001 & 0.10 & 21.19/0.5334 & 27.30/0.7504 & 26.97/0.7480 & 28.19/0.7738 & \textbf{29.34/0.8630} \\
0.001 & 0.25 & 22.20/0.6295 & 29.01/0.7847 & 29.26/0.8186 & 30.14/0.8251 & \textbf{34.68/0.9589} \\
\hline
0.002 & 0.04 & 18.54/0.3993 & 25.15/0.6474 & 24.79/0.6405 & 25.97/0.6697 & \textbf{28.21/0.8336} \\
0.002 & 0.10 & 20.02/0.4701 & 27.02/0.6873 & 26.59/0.7251 & 27.73/0.7461 & \textbf{29.34/0.8630} \\
0.002 & 0.25 & 20.50/0.5368 & 27.33/0.7031 & 28.06/0.7604 & 28.79/0.7628 & \textbf{34.65/0.9583} \\
\hline
0.004 & 0.04 & 17.45/0.3631 & 24.53/0.5984 & 24.59/0.6271 & 25.74/0.6528 & \textbf{28.21/0.8331} \\
0.004 & 0.10 & 18.70/0.3890 & 25.03/0.6196 & 25.94/0.6854 & 26.99/0.6989 & \textbf{29.32/0.8620} \\
0.004 & 0.25 & 18.73/0.4254 & 25.10/0.6522 & 26.49/0.6787 & 27.04/0.6789 & \textbf{34.54/0.9560} \\
\hline
\end{tabular}
}
\end{table}

%% file: Images/models_comparison.tex
\begin{figure}
  \centering
  \includegraphics[width=\linewidth]
  {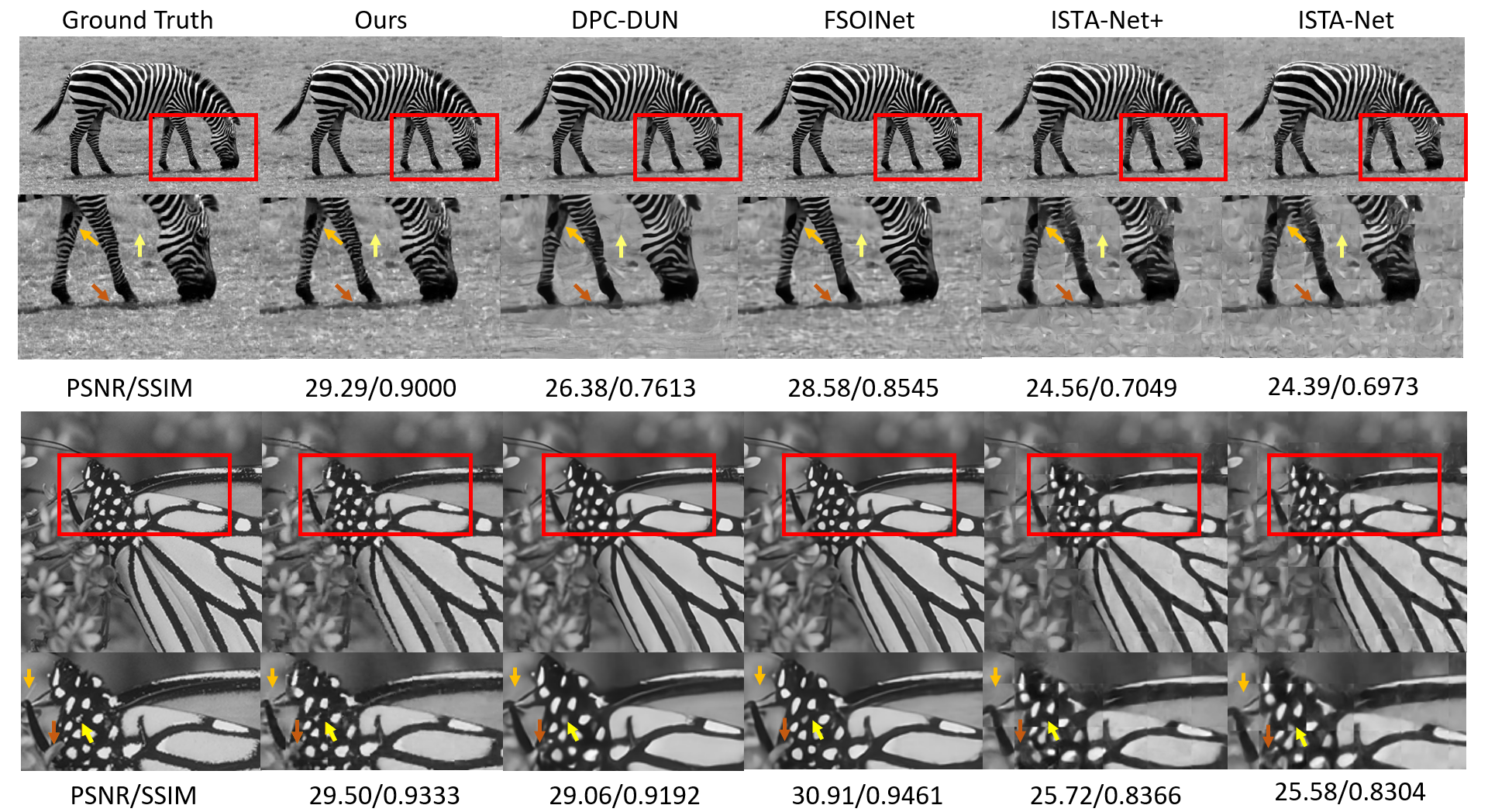}
  \caption{Performance comparison of different models at a compression ratio of 0.1 using images from the Set11 dataset.}
  \label{fig:modelcomparison}
\end{figure}

%% file: Images/meta_comparison.tex
\begin{figure}
  \centering
  \includegraphics[width=\linewidth]
  {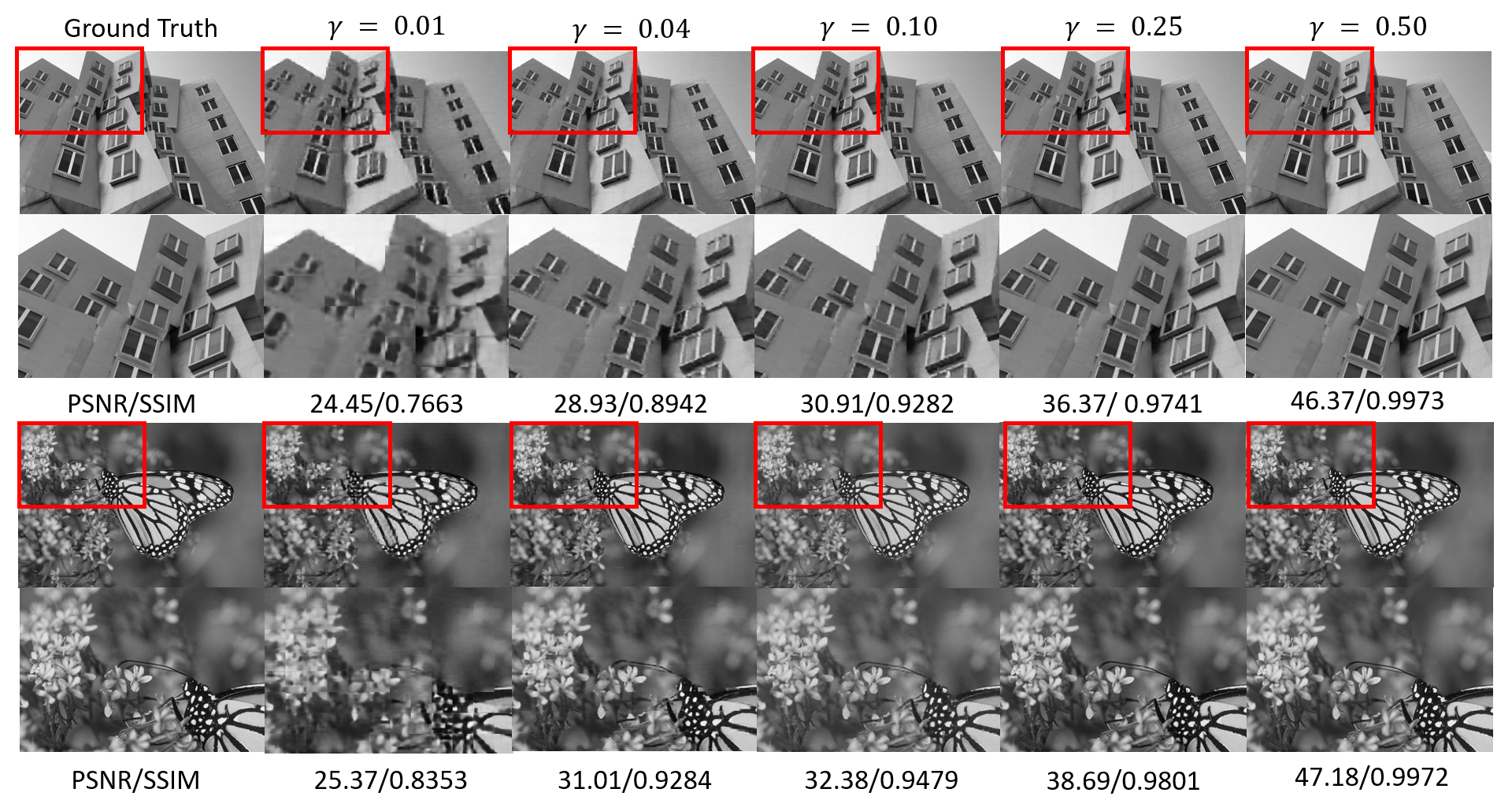}
  \caption{Visualizing the performance of our model across various compression ratios using images from the Set11 and Urban100 datasets}
  \label{fig:metacomparison}
\end{figure}

%% file: tables/ablation.tex
\begin{table}[!t]
    \caption{Ablation results}
    \begin{subtable}[t]{0.45\textwidth}
    \centering
    \caption{Comparing reconstruction performance with different transformer block counts on Set11 dataset at 0.25 compression ratio.}
    \label{tab:ablation}
        \begin{tabular}{cccc}
            \hline
            Encoder & Decoder & Parameters & PSNR/SSIM \\ 
            \hline
            32 & 32 & 847K &  35.28/0.967  \\
            20 & 20 & 542K &  33.47/0.956  \\
            16 & 16 & 440K & 36.11/0.970 \\
            12 & 12 & 339K &  36.71/0.976  \\
            8 & 8 & 237K &  33.56/0.952  \\
            4 & 4 & 136K &  33.77/0.960  \\
            \hline
        \end{tabular}
    \end{subtable}
    \hfill
    \begin{subtable}[t]{0.45\textwidth}
    \centering
    \caption{Comparison of reconstruction performance with and without the proposed embedding scheme $\mathbf{E}$ for 0.25 compression ratio.}
    \label{tab:withoutembedding}
        \begin{tabular}{ccc}
            \hline
            \multirow{2}{*}{Dataset} & \multicolumn{2}{c}{PSNR/SSIM}  \\ \cline{2-3}
                                     & No Embedding & $\mathbf{E}_{i,j}$          \\ \hline
            Set11                    & 21.72/0.592 & 36.00/0.970        \\
            BSD100                   & 23.60/0.557 & 34.79/0.959        \\
            Urban100                 & 19.95/0.486 & 32.53/0.963        \\ \hline
        \end{tabular}
    \end{subtable}
\end{table}

%% file: Sections/5_Discussion.tex
\section{Discussion}
\label{sec:discussion}

\textbf{Interpreting results:}
MOSAIC is able to improve the PSNR values with increments ranging from $+2 \ \text{dB}$ to $+7 \ \text{dB}$, compared to previous models.
Since PSNR focuses on the pixel-level differences between images, it can be inferred that our proposed method is effective in preserving fine-grained detail in the reconstruction.
Additionally, our model is able to achieve better performance in SSIM, implying that it performs well in terms of perceptual aspects and preserves the structural similarity. Although SOTA for higher compression ratios was already high (around $0.98$), MOSAIC was able to push the limit even further, demonstrating the effectiveness in the proposed method.

\textbf{Choice of sampling basis:}
The binary state of $\mathcal{H}$ is often seen as an inhibitor for CS reconstruction accuracy \cite{shen2022transcsnet}. However we have empirically disproven this idea through our results. Other popular choices for $\Phi$ could be a wavelet basis, or a DCT basis, depending on the specific application. Future work could explore these choices, and the effect of using structure-specific embeddings in the reconstruction models.

\textbf{Adaptive selection of measurement vectors:}
More recent work on deep learning methods for CS reconstruction attempt at selecting the sampling rows adaptively. In CASNet~\cite{chen2022casnet}, it samples each patch at an adaptive compression rate, so as to more efficiently allocate the given number of samples $m$ for an overall compression rate. Without such sophisticated adaptive sampling, MOSAIC outperforms CASNet, which further alludes to the effectiveness of the proposed reconstruction. (Table~\ref{tab:ssim-psnr-comp}, Set11 dataset~\cite{kulkarni2016reconnet})

\textbf{Limitations and future directions;}
Our proposed formalisms are derived for sampling matrices which have orthogonal, and separable transforms. Therefore, the formalism and approaches for training on random sampling matrices could be explored.
Furthermore, MOSAIC uses a patching approach for processing images, which creates the potential for blocking artifacts in reconstructions. Currently, there are models such as DPC-DUN~\cite{song2023dpcdun} which use overlapping blocks, although it is harder to reproduce in practical imaging systems. Therefore, it is promising to explore methods to overcome such qualitative defects without compromising applicability.

%% file: Sections/6_Conclusion.tex
\section{Conclusion}
\label{sec:conclusion}

In this work we introduce MOSAIC, a masked-encoding-based inverse solver for compressive sensing (CS). In MOSAIC we reformulate the CS inverse problem as a learnable measurement unmasking problem in a latent representation. Moreover, MOSAIC exploits the structure of a fixed sampling basis to embed measurements into a high-dimensional representation before encoding. Our results suggest that the proposed embedding, along with masked optimisation, allows efficient reconstructions. To this end, MOSAIC achieves the state of the art across a wide range of compression ratios on a variety of benchmarks, even outperforming methods with learnable sampling matrices. We believe that the strong inductive bias given by explicitly using the structure of the measurement vectors, accounts for such improvements. Thus, we demonstrate the possibility of learning a general inverse mapping from a random selection of samples, for CS reconstruction through a given basis.

%% file: Sections/7_Supplementary.tex
\section{Supplementary}

\subsection{More details in the experimental setup}

\textbf{Distributed training}
The base model was trained on two Nvidia Quadro RTX 6000 GPUs (24GB RAM), while ablations, and extensive experiments were run on four NVIDIA A100-SXM4-40GB GPUs (32GB RAM). The accelerate library~\cite{huggingface_accelerate} was employed to enable the codes to be trained on any distributed configuration.

\textbf{Training images}
The proposed architecture was designed for reconstructing an image of a fixed size. For our experiments, the fixed size of an input $\mathbf{X}$ was (32,32). In order to accommodate images with other sizes and shapes, images were zero-padded and sliced to generate non-overlapping patches as a pre-processing step during validation and testing~\cite{istanet2018,istanetplus2021}. The training dataset CIFAR10~\cite{Krizhevsky2009LearningMLCIFAR10}, contained images of same size. Datasets BSD500~\cite{bsd100} (used for fine-tuning), and BSD100~\cite{bsd100}, Set11~\cite{kulkarni2016reconnet}, Urban100~\cite{Huang-CVPR-2015} contained images with different (larger) dimensions. This required the padding and patching process as pre-processing step, and the stitching of these patches together as post-processing.
    
\textbf{Training and inference}
We used the functions provided by torch-metrics library with default parameters to calculate the PSNR, SSIM metrics. Validation accuracy (PSNR) curves for different compression ratios are shown in figure~\ref{fig:typicaltrainingcurves}. We noticed that MOSAIC converges faster for low compression ratios. Therefore, it is possible that some models would require more training than the recommended 500 epochs with changes in training data.\\
We analysed existing literature, to find the most commonly inferred test datasets. Our trained model was then inferred on these datasets, and the results were reported in the main results table. Our complete analysis is shown in Table~\ref{all-results}, where `-' indicates unavailable model weights, which require separate training, and `*' indicates results that have been reproduced with publicly available model weights.


\subsection{Extended Quantitative Results}

\textbf{Random sampling}
In contrast to preceding approaches, we propose a reconstruction model that is applicable for any random choice of measurements. To inspect the consistency of reconstruction performance, we infer our model on multiple random samples on test data.

We reported the mean and standard deviation values of the inferences in Table~\ref{all-results}.
where the inferences are repeated for 10 random instances of the compressed sampling. The random seed was changed each instance, which created a different sampling of $\mathbf{Y}$. 
The low standard deviation indicates the consistency of the proposed reconstruction for any random sample.

\textbf{Robustness to noise}
We have reported on the ability of MOSAIC to reconstruct images, even in instances the input image is corrupted with Gaussian noise. This was being consistent with previous work~\cite{shen2022transcsnet}, in selecting the compression ratios ($\gamma$), and the standard deviation ($\sigma$) for the noise. Here we extend the comparison to a wider range of $\sigma$ and $\gamma$ in Table~\ref{table:noise_extended_psnr}, to analyse the effect of higher levels of corruption in inputs for MOSAIC.

\textbf{Evaluating across compression ratios}\\
In general, a MOSAIC model trained specifically for a given compression ratio $\gamma$ outperformed other versions of the model in terms of validation and test metrics. Interestingly, similar to the empirical observations made in~\cite{mildenhall2021nerf}, we noticed that MOSAIC trained on compression ratios that are binary fractions, ($\gamma = 0.125 = 1/4,\ \gamma = 0.03125 = 1/32$) slightly outperforms models trained on other relatively closer $\gamma$.
Maintaining consistency, we have not considered these results in our comparisons in Table~\ref{all-results}, although we report this observation in Table~\ref{tab:crosscompratios}.

\input{SupplementaryImages/TrainingCurves}

\input{SupplementaryTables/allexistingrelatedwork}

\newpage
\input{SupplementaryTables/crosscompratios}

\input{SupplementaryTables/noiceextended}


\newpage
\subsection{Extended Qualitative Results}

\textbf{Comparison with other models}
We have reported the compression ratio $\gamma = 0.1$ qualitative comparison in the main paper. This is since the differences among models are visually apparent at such low compression ratios. For higher $\gamma = 0.25, \ 0.5$, although there are differences in the PSNR/SSIM metrics, these differences are not prominent in the visualisation. This is evident from the comparison in Figures
\ref{fig:modelcomparison_highercomps_0.25},\ref{fig:modelcomparison_highercomps_0.5} For lower compression ratios, we use $\gamma = 0.1$ (Figure~\ref{fig:modelcomparison_highercomps_0.1}) since it has the lowest compression ratio with common published weights among related work. We observe better sharpness and reconstruction of fine details in our method. Additionally, we include a comparison of 3-channel images (Figure \ref{fig:modelcomparison_3channel}), following related work\cite{istanetplus2021, zhang2020ampnet, song2023dpcdun} in which only the Lumina(Y) channel is processed through the CS algorithm. Please zoom in for a clear comparison.

\textbf{Visualising reconstructions of corrupted inputs}\\
We further evaluate the robustness to noise of MOSAIC, compared to recent work, in a qualitative sense for a higher noise $\sigma = \ 0.1$, and compression ratios $\gamma = 0.1, \ 0.25$ in Figure~\ref{fig:noisedinputs_reconstruction}. This demonstrates that although MOSAIC is resilient to low noise levels, it is less robust in higher $\sigma$. Although our main goal is not to achieve noise robustness, future work could explore on how to improve noise-robustness through incorporating the MOSAIC architecture with noise statistics .


\input{SupplementaryImages/modelcomparison_highercomps}
\input{SupplementaryImages/noise_images}

\newpage
\subsection{Extended Ablations}
\textbf{Choice of sampling matrix}\\
The Walsh-Hadamard matrix $\mathbf{\Phi} = \mathcal{H}_{(1,-1)}$ which was used in MOSAIC has binary elements, with \{1, -1\} entries. This choice was motivated by it's orthogonal property, and
empirical evidence of better learning (Table~\ref{tab:choiceofhadamard}); when compared to the non-orthogonal Hadamard matrix $\mathcal{H}_{(1,0)} $ with \{1, 0\} elements. Considering a practical implementation with a sampling scheme employing $\mathcal{H}_{(1,0)}$, we note the relation between the sampled measurements from the ith row of both matrices. Specifically, $p_{i} = \mathcal{H}_{(1,0), \ i} \  \mathbf{x}$ is related to the measurement $q_{i} = \mathcal{H}_{(1,-1), \ i} \ \mathbf{x}$, through a measurement $\mathbf{1}^{T} \mathbf{x}$ as follows.\\
\begin{align*}
    p_{i} & =  (q_{i} + \mathbf{1}^{T}\mathbf{x} )/2 \\
\end{align*}
Hence, provided that the first row of $\mathcal{H}_{(1,0)}$ (which is $\mathbf{1}^{T}$), is always selected as a measurement vector, one could use the above equation to create a mapping between MOSAIC's proposed reconstruction algorithm, and a more practically feasible CS scheme. Noting that the above relation would not hold exactly under noisy measurements, one should consider analysing the noise distributions in measurement space. 

\textbf{Parameter counts}\\
Following recent work, we compare the parameter count of MOSAIC with other models at different compression ratios in figure \ref{fig:psnr_params_extended}. The values for other models are reported from previous work~\cite{shen2022transcsnet}, and publicly available inference codes.

\input{SupplementaryTables/hadamardchoice}
\input{SupplementaryImages/psnr_vs_params}

%% file: SupplementaryImages/TrainingCurves.tex
\begin{figure}[htbp]
  \centering
  \includegraphics[width=0.7\linewidth]{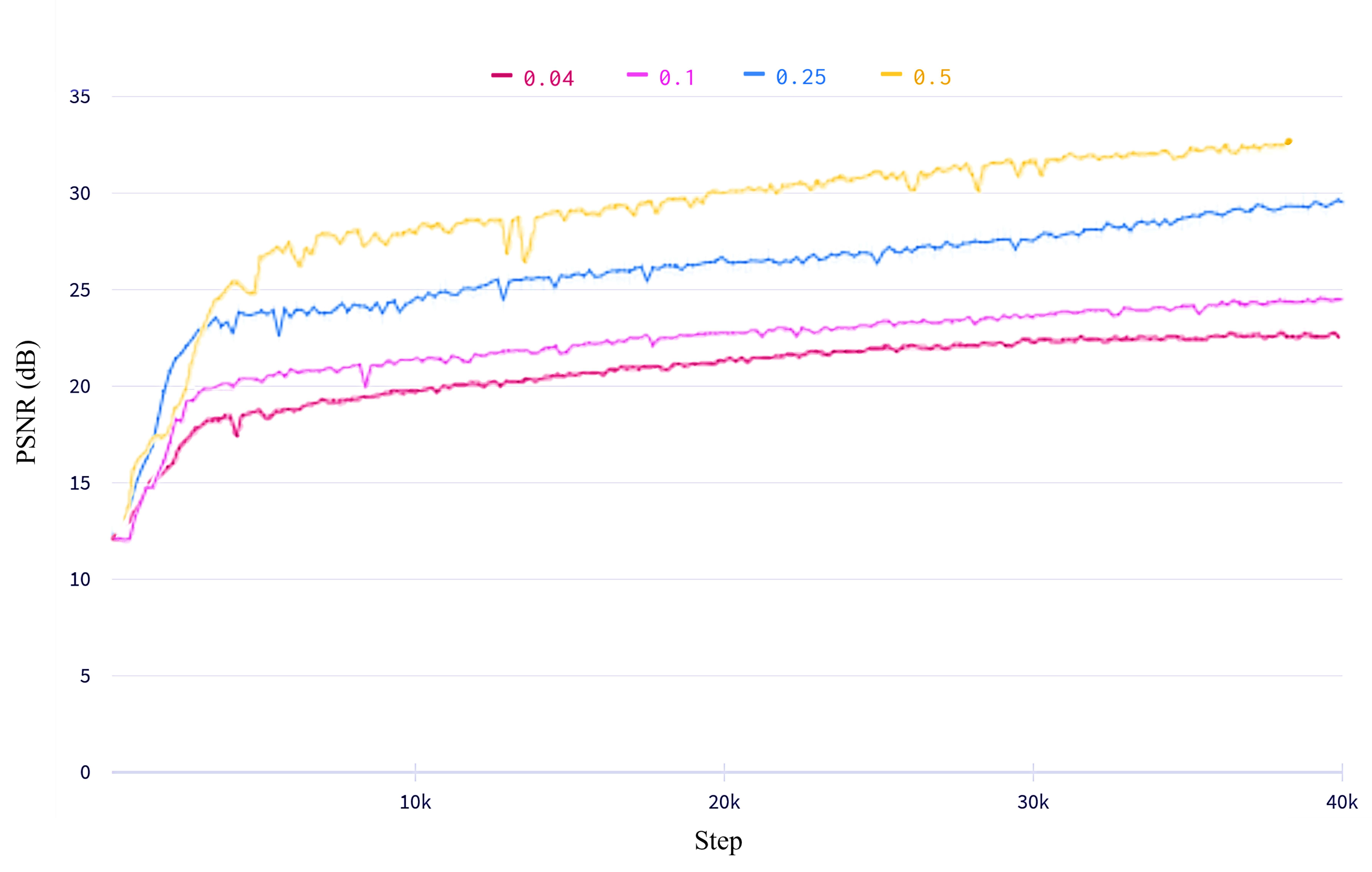}
  \caption{A model showing the validation PSNR curves for the $\gamma = 0.5, 0.25, 0.1 \text{ and }0.04 $ MOSAIC models}
  \label{fig:typicaltrainingcurves}
\end{figure}

%% file: SupplementaryTables/allexistingrelatedwork.tex
\begin{table}
  \caption{PSNR/SSIM metric comparisons of different models on test datasets}
  \label{all-results}
  \centering
  \resizebox{\textwidth}{!}{\begin{tabular}{clccccc}
    \toprule
    & & \multicolumn{5}{c}{PSNR/SSIM for sampling ratios}\\
    \cmidrule{3-7}
    Dataset & Method & 0.01 & 0.04 & 0.1 & 0.25 & 0.5 \\
    \toprule

    \multirow{9}{*}{\rotatebox[origin=c]{90}{Set5}} & CSNet+~\cite{shi2019imageCNN} & - & 28.22/0.809 & 32.22/0.911 & 36.35/0.958 & 41.61/0.983 \\
     & ISTA-Net++~\cite{istanetplus2021} & - & - & - & - & - \\
     & AMP-Net~\cite{zhang2020ampnet} & - & 27.71/0.811 & 31.79/0.898 & 36.68/0.953 & 41.52/0.979 \\
     & TransCS~\cite{shen2022transcsnet} & - & 29.02/0.842 & \textbf{33.56/0.924} & 38.26/0.965 & 43.42/0.985 \\
     & FSOINet~\cite{chen2022fsoinet} & 23.18/0.639* & 28.60/0.846* & 33.04/0.924* & 37.53/0.963* & 42.65/0.984* \\
     & CASNet~\cite{chen2022casnet} & - & - & - & - & - \\ 
     & DPCDUN~\cite{song2023dpcdun} & - & - & 31.12/0.893* & 35.96/0.9471* & 41.07/0.978* \\
     & \textbf{MOSAIC (mean)} & \textbf{25.95/0.697} & \textbf{31.15/0.899} & 32.80/0.920 & \textbf{38.34/0.969} & \textbf{47.21/0.996} \\
    & \textbf{MOSAIC std} & 0.0456/0.0023 & 0.0738/0.0011 & 0.08/0.0004 & 0.0000/0.0426 & 0.2804/0.0004\\ \midrule

    \multirow{9}{*}{\rotatebox[origin=c]{90}{Set11}} & CSNet+~\cite{shi2019imageCNN} & 20.67/0.541 & 24.83/0.784 & 28.34/0.858 & 33.34/0.939 & 38.47/0.979 \\
     & ISTA-Net++~\cite{istanetplus2021} & 18.29/0.459 & 22.91/0.679 & 27.54/0.835 & 32.70/0.927 & 37.47/0.967 \\
     & AMP-Net~\cite{zhang2020ampnet} & 20.20/0.558 & 25.26/0.772 & 29.40/0.878 & 34.63/0.948 & 40.33/0.980 \\
     & TransCS~\cite{shen2022transcsnet} & - & - & - & - & - \\
     & FSOINet~\cite{chen2022fsoinet} & 21.73/0.594 & 26.37/0.812* & \textbf{30.44}/0.902 & 35.80/0.960* & 41.08/0.983 \\
     & CASNet~\cite{chen2022casnet} & 21.97/0.614 & 26.41/0.815 & 30.36/0.901 & 35.67/0.959 & 40.93/0.983 \\ 
     & DPC-DUN~\cite{song2023dpcdun} & - & - & 29.40/0.880 & 34.69/0.948 & 39.84/0.978 \\
     & \textbf{MOSAIC (mean)} & \textbf{23.68}/\textbf{0.710} & \textbf{28.28}/\textbf{0.871} & 29.93/\textbf{0.905} & \textbf{36.02}/\textbf{0.970} & \textbf{45.46}/\textbf{0.997} \\
    & \textbf{MOSAIC std} & 0.0375/0.0007 & 0.0546/0.0009 & 0.0236/0.0003 & 0.036/0.0 & 0.2812/0.0 \\ \midrule


    \multirow{9}{*}{\rotatebox[origin=c]{90}{BSD100}}  & CSNet+~\cite{shi2019imageCNN} & - & 25.79/0.674 & 28.29/0.796 & 31.50/0.899  & 36.35/0.964 \\
     & ISTA-Net++~\cite{istanetplus2021} & 19.98/0.449 & 23.26/0.587 & 25.86/0.717 & 29.48/0.854 & 33.97/0.941 \\
     & AMP-Net~\cite{zhang2020ampnet} & 23.45/0.554 & 26.35/0.681 & 28.87/0.792 & 32.69/0.903 & 37.71/0.967 \\
     & TransCS~\cite{shen2022transcsnet} & - & 26.25/0.689 & 28.79/0.806 & 32.66/0.910  & 37.70/0.969 \\
     & FSOINet~\cite{chen2022fsoinet} & 22.76/0.528* & 25.44/0.686* & 28.01/0.806* & 31.90/0.911* & 37.02/0.970* \\
     & CASNet~\cite{chen2022casnet} & - & - & - & - & - \\ 
     & DPCDUN~\cite{song2023dpcdun} & - & - & 26.54/0.746* & 30.43/0.874* & 35.32/0.954* \\
     & \textbf{MOSAIC (mean)} & \textbf{24.59}/\textbf{0.577} & \textbf{28.12}/\textbf{0.832} & \textbf{29.33}/\textbf{0.866} & \textbf{34.63}/\textbf{0.955} & \textbf{45.07}/\textbf{0.996} \\
    & \textbf{MOSAIC std} & 0.008/0.0 & 0.0068/0.0003 & 0.0066/0.0 & 0.0074/0.0 & 0.0319/0.0 \\ \midrule

    \multirow{9}{*}{\rotatebox[origin=c]{90}{Urban100}}  & CSNet+~\cite{shi2019imageCNN} & - & 21.17/0.611 & 24.01/0.783 & 27.68/0.897 & 33.32/0.966 \\
    & ISTA-Net++\cite{istanetplus2021} & 17.48/0.416 & 20.96/0.599 & 24.66/0.762 & 29.51/0.894 & 34.41/0.957 \\
    & AMP-Net~\cite{zhang2020ampnet} & 20.90/0.533 & 24.15/0.703 & 27.38/0.827 & 32.19/0.926 & 37.51/0.973 \\
    & TransCS~\cite{shen2022transcsnet} & - & 23.23/0.702 & 26.73/0.842 & 31.72/0.933  & 37.20/0.976 \\
    & FSOINet~\cite{chen2022fsoinet} & 19.87/0.522 & 22.71/0.704* & \textbf{27.53}/0.863 & 32.17/0.950* & 37.80/0.978 \\
    & CASNet~\cite{chen2022casnet} & - & - & - & - & - \\
    & DPC-DUN~\cite{song2023dpcdun} & - & - & 26.99/0.835 & 32.36/0.932 & 37.52/0.974 \\
    & \textbf{MOSAIC (mean)}  & \textbf{21.35}/\textbf{0.571} & \textbf{25.20}/\textbf{0.830} & 26.72/\textbf{0.875} & \textbf{32.43}/\textbf{0.961} & \textbf{42.41}/\textbf{0.996} \\
    & \textbf{MOSAIC std}  & 0.0057/0.0003 & 0.0058/0.0005 & 0.0048/0.0003 & 0.0079/0.0 & 0.0741/0.0 \\
    \bottomrule
  \end{tabular}}
\end{table}

%% file: SupplementaryTables/crosscompratios.tex
\begin{table}[t]
    \centering
    \caption{MOSAIC trained on $\gamma = 0.125$, and inferred on samples of $\gamma = 0.1$}
    \label{tab:crosscompratios}
    \begin{tabular}{cccc}
        \hline
        \multirow{2}{*}{Dataset} & \multicolumn{3}{c}{PSNR/SSIM}  \\ \cline{2-4}
                     & MOSAIC (0.1) & MOSAIC (0.125) & Recent best         \\ \hline
        Set11        & 27.93/0.905 & \textbf{30.43/0.905} & 30.44 (FSOINet)/0.902 (FSOINet)         \\
        BSD100       & 29.33/\textbf{0.886} & \textbf{29.93}/0.884 & 28.87 (AmpNet)/0.806 (TransCS)  \\
        Urban100     & 26.72/0.875 & \textbf{27.53/0.891} & \textbf{27.53} (FSOINet)/0.863 (FSOINet)          \\ \hline
    \end{tabular}
    \vspace{3mm}
\end{table}

%% file: SupplementaryTables/noiceextended.tex
\begin{table}[!t]
\vspace{-10pt}
\centering
\caption{Evaluation of robustness to noise during inference on BSD100 dataset.}
\label{table:noise_extended_psnr}
\begin{tabular}{cccccc}
\hline
    \multirow{2}{*}{$\sigma$} & \multicolumn{5}{c}{PSNR for sampling ratios and reduction in accuracy compared to $\sigma = 0$}\\
    \cmidrule{2-6}
    & 0.01 & 0.04 & 0.1 & 0.25 & 0.5 \\
    \hline
    0.001 & 24.58 ($\downarrow 0.01$) & 28.11 ($\downarrow 0.01$) & 29.32 ($\downarrow 0.01$) & 34.63 ($\downarrow 0.00$) & 44.88 ($\downarrow 0.19$) \\
    0.002 & 24.58 ($\downarrow 0.01$) & 28.11 ($\downarrow 0.01$) & 29.32 ($\downarrow 0.01$) & 34.60 ($\downarrow 0.03$) & 44.43 ($\downarrow 0.64$) \\
    0.004 & 24.58 ($\downarrow 0.01$) & 28.10 ($\downarrow 0.02$) & 29.30 ($\downarrow 0.03$) & 34.48 ($\downarrow 0.15$) & 43.06 ($\downarrow 2.01$) \\
    0.01 & 24.57 ($\downarrow 0.02$) & 28.04 ($\downarrow 0.08$) & 29.19 ($\downarrow 0.14$) & 33.76 ($\downarrow 0.87$) & 38.69 ($\downarrow 6.38$) \\
    0.02 & 24.55 ($\downarrow 0.04$) & 27.83 ($\downarrow 0.29$) & 28.79 ($\downarrow 0.54$) & 31.90 ($\downarrow 1.73$) & 33.63 ($\downarrow 11.44$) \\
    0.04 & 24.46 ($\downarrow 0.13$) & 27.02 ($\downarrow 1.1$) & 27.47 ($\downarrow 1.86$) & 28.15 ($\downarrow 6.48$) & 27.89 ($\downarrow 17.18$) \\
    0.1 & 23.79 ($\downarrow 0.80$) & 23.38 ($\downarrow 4.74$) & 22.95 ($\downarrow 6.38$) & 21.23 ($\downarrow 13.40$) & 20.06 ($\downarrow 27.01$) \\
    0.2 & 21.77 ($\downarrow 2.82$) & 18.79 ($\downarrow 9.33$) & 17.92 ($\downarrow 11.41$) & 15.58 ($\downarrow 19.05$) & 14.28 ($\downarrow 30.79$) \\
    0.4 & 17.92 ($\downarrow 6.67$) & 13.87 ($\downarrow 14.25$) & 12.15 ($\downarrow 17.18$) & 10.34 ($\downarrow 24.29$) & 8.88 ($\downarrow 36.19$)  \\
    \hline
\end{tabular}
\vspace{2mm}
\end{table}

\begin{table}[!t]
\vspace{-10pt}
\centering
\caption{Evaluation of robustness to noise during inference on BSD100 dataset.}
\label{table:noise_extended_ssim}
\begin{tabular}{cccccc}
\hline
    \multirow{2}{*}{$\sigma$} & \multicolumn{5}{c}{SSIM for sampling ratios and reduction in accuracy compared to $\sigma = 0$}\\
    \cmidrule{2-6}
    & 0.01 & 0.04 & 0.1 & 0.25 & 0.5 \\
    \hline
    0.001 & 0.577 ($\downarrow 0.000$) & 0.832 ($\downarrow 0.000$) & 0.866 ($\downarrow 0.000$) & 0.955 ($\downarrow 0.000$) & 0.994 ($\downarrow 0.002$) \\
    0.002 & 0.577 ($\downarrow 0.000$) & 0.832 ($\downarrow 0.000$) & 0.866 ($\downarrow 0.000$) & 0.954 ($\downarrow 0.001$) & 0.993 ($\downarrow 0.003$) \\
    0.004 & 0.577 ($\downarrow 0.000$) & 0.832 ($\downarrow 0.000$) & 0.865 ($\downarrow 0.001$) & 0.951 ($\downarrow 0.004$) & 0.988 ($\downarrow 0.008$) \\
    0.01 & 0.576 ($\downarrow 0.001$) & 0.828 ($\downarrow 0.004$) & 0.858 ($\downarrow 0.008$) & 0.933 ($\downarrow 0.022$) & 0.956 ($\downarrow 0.040$) \\
    0.02 & 0.575 ($\downarrow 0.002$) & 0.814 ($\downarrow 0.018$) & 0.833 ($\downarrow 0.033$) & 0.876 ($\downarrow 0.079$) & 0.874 ($\downarrow 0.122$) \\
    0.04 & 0.569 ($\downarrow 0.008$) & 0.761 ($\downarrow 0.071$) & 0.754 ($\downarrow 0.112$) & 0.734 ($\downarrow 0.221$) & 0.707 ($\downarrow 0.289$) \\
    0.1 & 0.531 ($\downarrow 0.046$) & 0.552 ($\downarrow 0.280$) & 0.527 ($\downarrow 0.339$) & 0.451 ($\downarrow 0.504$) & 0.411 ($\downarrow 0.585$) \\
    0.2 & 0.433 ($\downarrow 0.144$) & 0.352 ($\downarrow 0.480$) & 0.328 ($\downarrow 0.538$) & 0.242 ($\downarrow 0.713$) & 0.219 ($\downarrow 0.777$) \\
    0.4 & 0.279 ($\downarrow 0.298$) & 0.184 ($\downarrow 0.648$) & 0.199 ($\downarrow 0.667$) & 0.110 ($\downarrow 0.845$) &0.108 ($\downarrow 0.888$)  \\
    \hline
\end{tabular}
\end{table}

%% file: SupplementaryImages/modelcomparison_highercomps.tex
\begin{figure}[htbp]
  \centering
  \includegraphics[width=\linewidth]
  {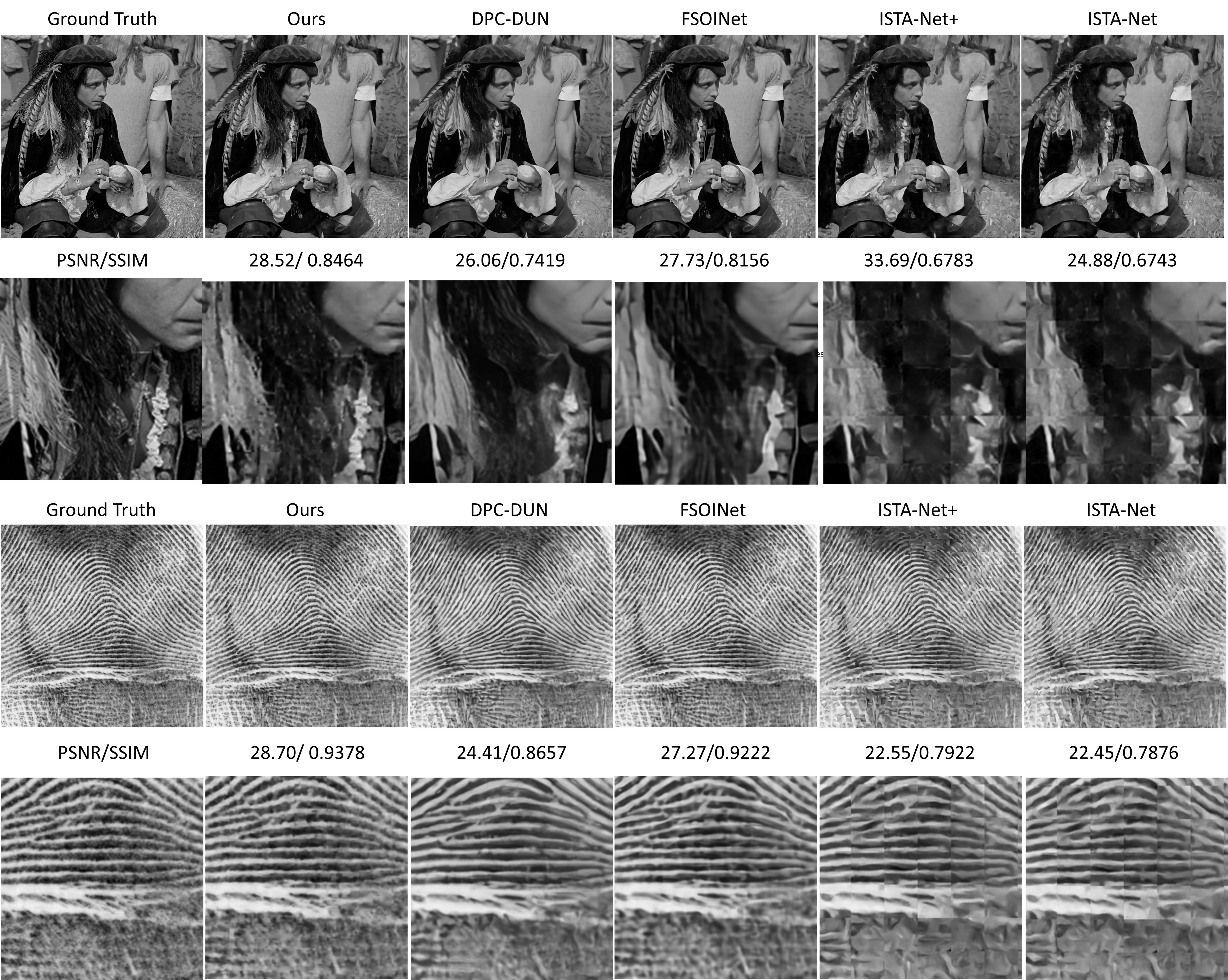}
  \caption{Comparing MOSAIC to other models, showing comparable visual differences for $\gamma = 0.1$. Ground-truth images are from Set14 and Set11 datasets respectively.}
  \label{fig:modelcomparison_highercomps_0.1}
\end{figure}

\begin{figure}[htbp]
  \centering
  \includegraphics[width=\linewidth]
  {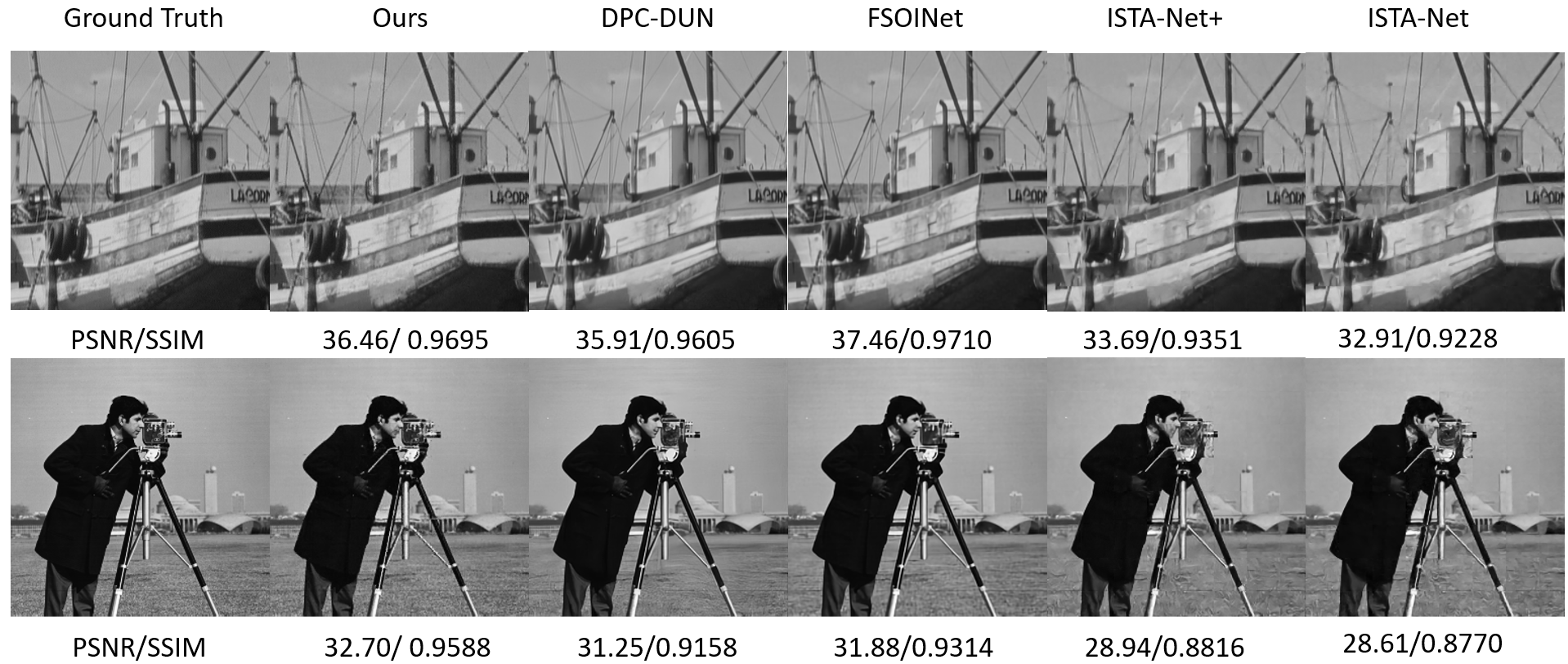}
  \caption{Comparing MOSAIC to other models, showing the almost negligeble visual differences for $\gamma = 0.25$. Ground-truth images are  from the Set11 dataset.}
  \label{fig:modelcomparison_highercomps_0.25}
\end{figure}

\begin{figure}[htbp]
  \centering
  \includegraphics[width=\linewidth]
  {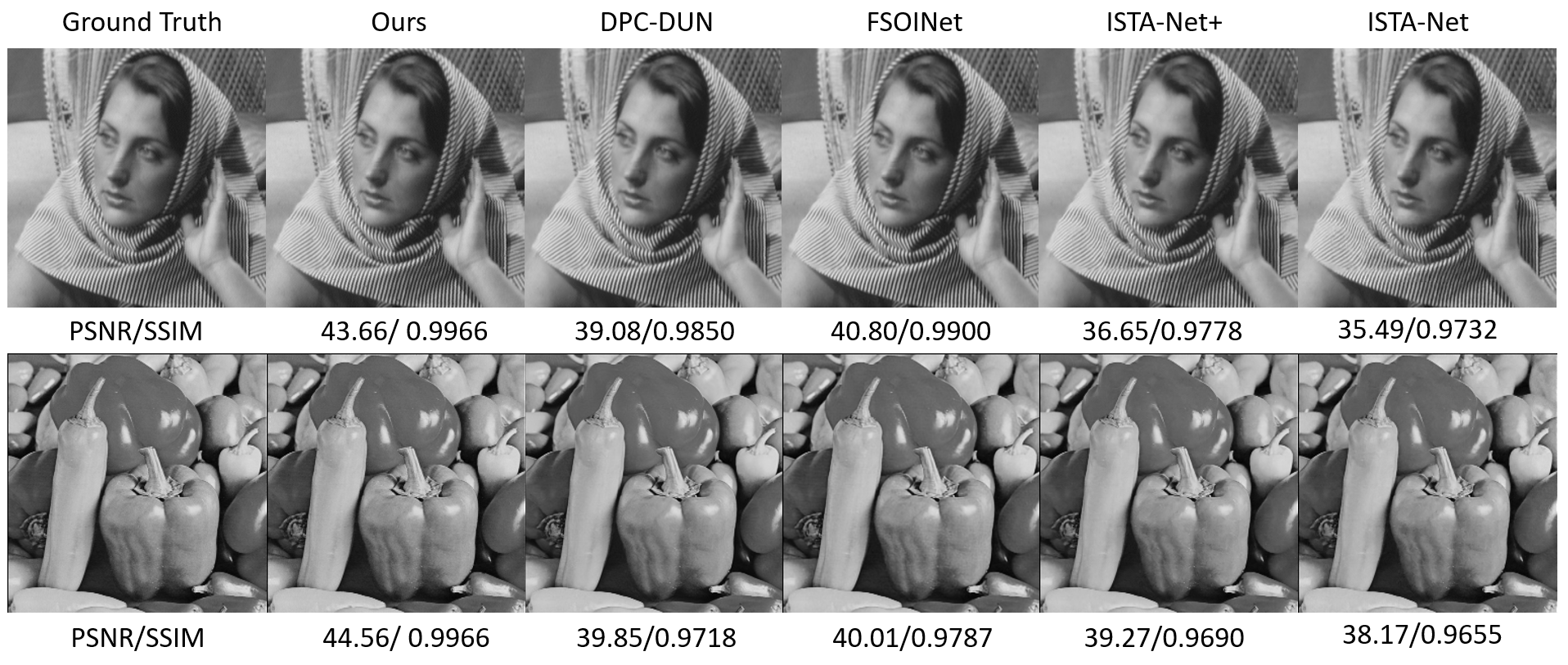}
  \caption{Comparing MOSAIC to other models, showing the almost negligeble visual differences for $\gamma = \ 0.5$. Ground-truth images are from the Set11 dataset.}
  \label{fig:modelcomparison_highercomps_0.5}
\end{figure}

\begin{figure}[htbp]
  \centering
  \includegraphics[width=\linewidth]
  {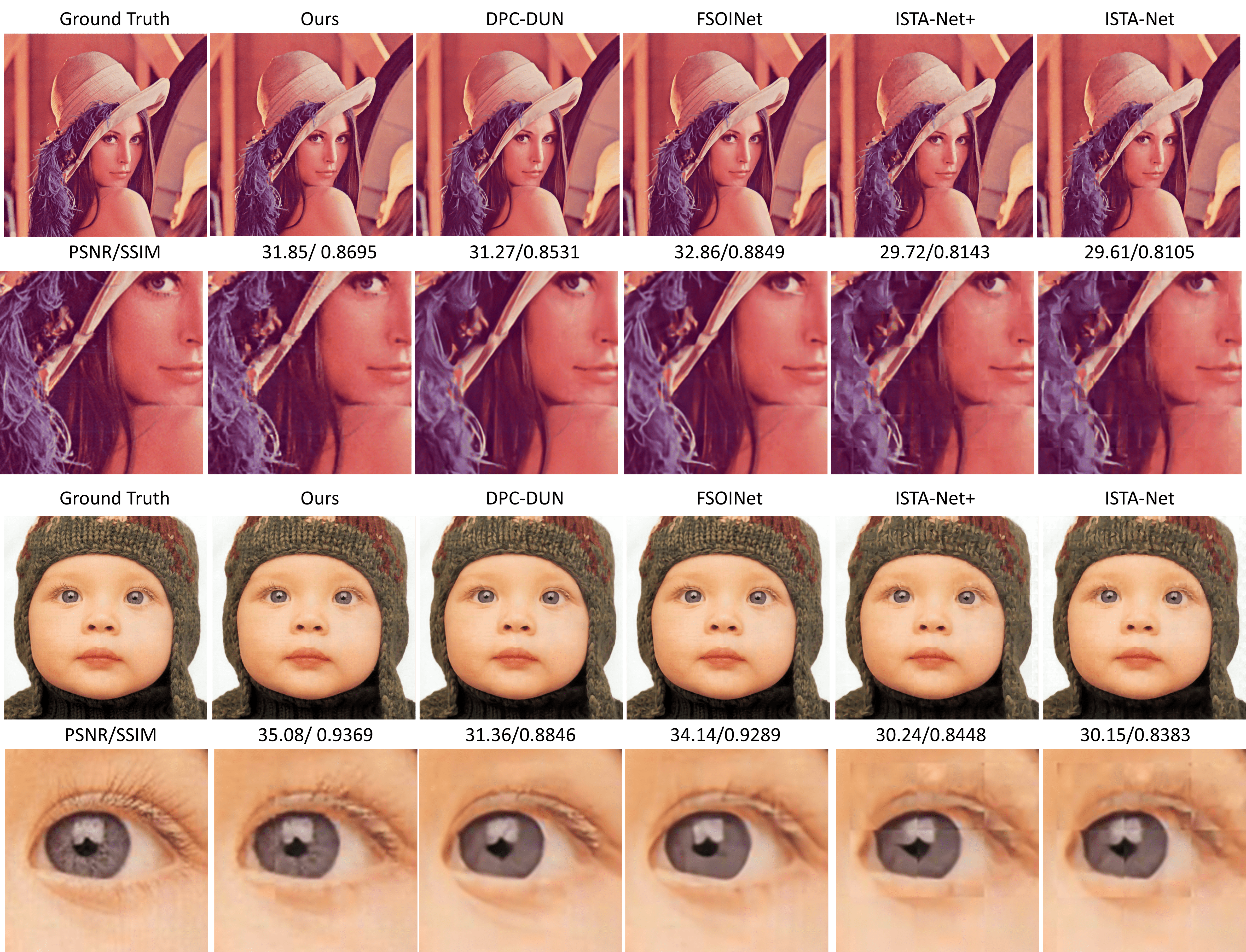}
  \caption{Comparing MOSAIC to other models,depicting visual differences for $\gamma = \ 0.1$ in a 3-channel visualisation. Ground-truth images are taken from Set5 and Set14 datasets respectively.}
  \label{fig:modelcomparison_3channel}
\end{figure}

%% file: SupplementaryImages/noise_images.tex
\begin{figure}[htbp]
  \centering
  \includegraphics[width=\linewidth]
  {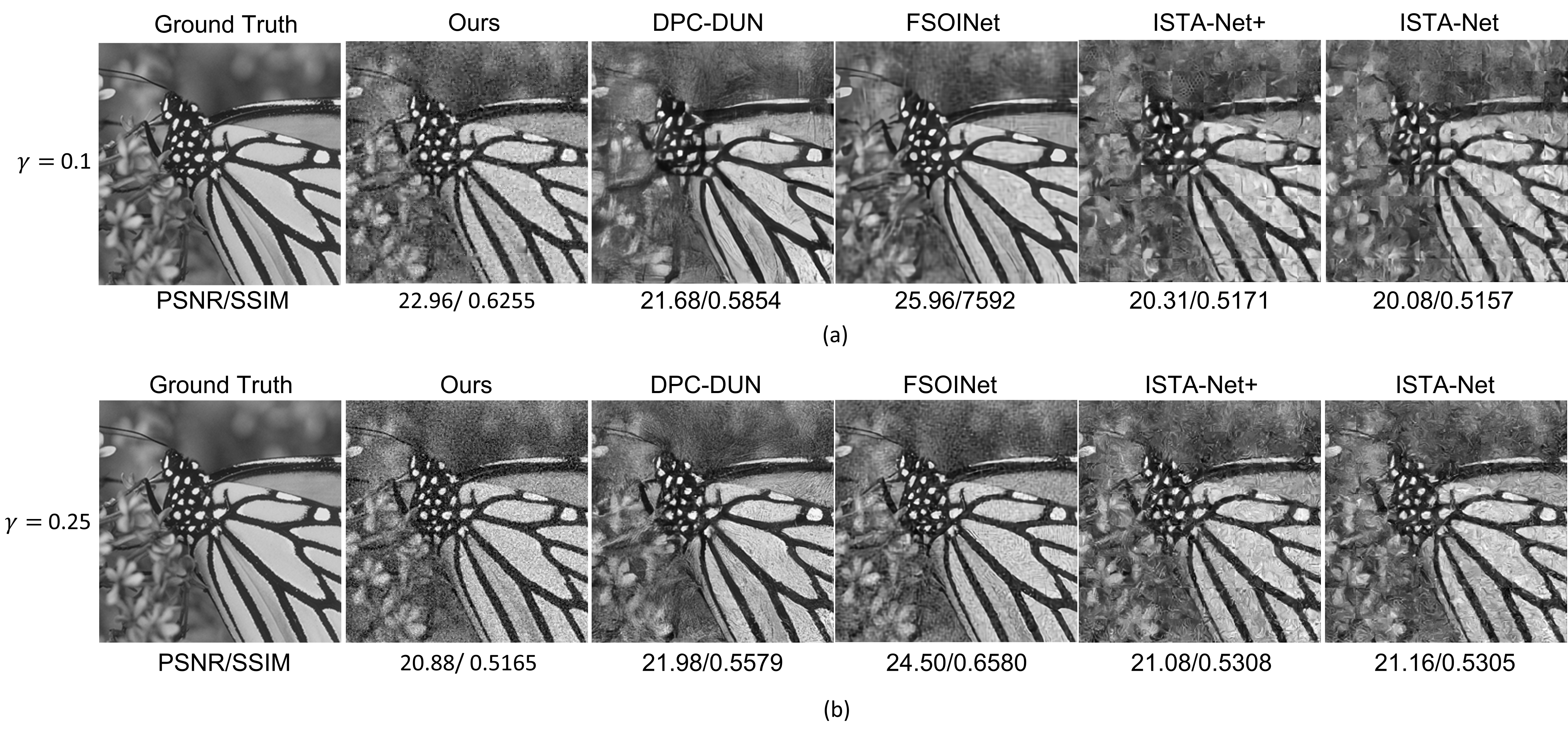}
  \caption{Qualitative evaluation of reconstruction models for a Gaussian-noise corrupted image at $\sigma = 0.1$.}
  \label{fig:noisedinputs_reconstruction}
\end{figure}

%% file: SupplementaryTables/hadamardchoice.tex
\begin{table}[t]
    \centering
    \caption{Comparison of $\mathcal{H}_{(1,0)}$ with $\mathcal{H}_{(1,-1)}$ \\ 
 at $\gamma = 0.25$ }
    \label{tab:choiceofhadamard}
        \begin{tabular}{ccc}
            \hline 
            $\mathbf{\Phi}$ & val. PSNR & val. SSIM \\
            \hline
            $\mathcal{H}_{(1,0)}$    & 12.4 &   0.14      \\
            $\mathcal{H}_{(1,-1)}$   & 33.0 &   0.96      \\ \hline
        \end{tabular}
\end{table}

%% file: SupplementaryImages/psnr_vs_params.tex
\begin{figure}[htbp]
  \centering
  \includegraphics[width=\linewidth]
  {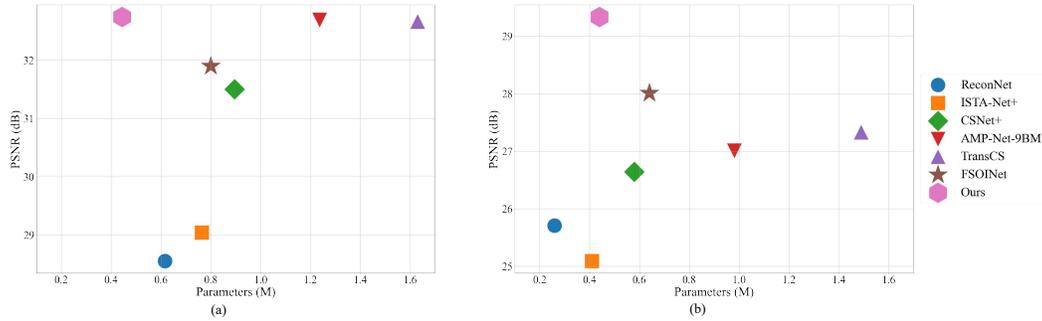}
  \caption{Comparison of PSNR with the number of parameters: for $\gamma = 0.1 \text{ and } \gamma = 0.25$}
  \label{fig:psnr_params_extended}
\end{figure}